\documentclass[letterpaper]{article} 
\usepackage{aaai23}  
\usepackage{times}  
\usepackage{helvet}  
\usepackage{courier}  
\usepackage[hyphens]{url}  
\usepackage{graphicx} 
\usepackage{natbib}  
\usepackage{caption} 
\usepackage{booktabs}
\usepackage{algorithm}
\usepackage{algorithmic}
\usepackage{comment}

\usepackage{newfloat}
\usepackage{listings}
\DeclareCaptionStyle{ruled}{labelfont=normalfont,labelsep=colon,strut=off} 
\floatstyle{ruled}
\newfloat{listing}{tb}{lst}{}
\floatname{listing}{Listing}
\usepackage{amsmath}
\usepackage{amsfonts}
\usepackage{tabularx}
\usepackage{bm}

\title{Learning Joint Representation of Human Motion and Language}
\author {
    Jihoon Kim,\textsuperscript{\rm 1 *}
    Youngjae Yu, \textsuperscript{\rm 2 *}
    Seungyoun Shin, \textsuperscript{\rm 1}
    Taehyun Byun, \textsuperscript{\rm 1}
    Sungjoon Choi \textsuperscript{\rm 1}
}
\affiliations {
    \textsuperscript{\rm 1} Korea University\\
    \textsuperscript{\rm 2}  Allen Institute for AI\\
    jihoon-kim@korea.ac.kr, youngjaey@allenai.org, 2022021568@korea.ac.kr, taehyun-byun@korea.ac.kr, sungjoon-choi@korea.ac.kr
}

\begin{document}

\maketitle

\begin{abstract}
In this work, we present MoLang (a Motion-Language connecting model) for learning joint representation of human motion and language, leveraging both unpaired and paired datasets of motion and language modalities. To this end, we propose a motion-language model with contrastive learning, empowering our model to learn better generalizable representations of the human motion domain. Empirical results show that our model learns strong representations of human motion data through navigating language modality. Our proposed method is able to perform both action recognition and motion retrieval tasks with a single model where it outperforms state-of-the-art approaches on a number of action recognition benchmarks. 
\end{abstract}

\section{Introduction}
As humans, we can easily generalize and imitate human motion and even understand the contexts of human motion in the wild. However, it will be a long and detour way for an autonomous agent (e.g., a humanoid robot) to learn how to understand motions from limited motion capture data. Nevertheless, a humanoid capable of representing and expressing its detailed motion holds practical promise. With the advent of scalable Transformer~\cite{vaswani2017attention}, significant progress has been made in joining multiple modalities such as image-text~\cite{zhang2020contrastive,radford2021learning,jia2021scaling,tan2019lxmert,li2019visualbert,lu2019vilbert,chen2019uniter,yu2020ernie}, video/audio-text~\cite{rouditchenko2020avlnet,wang2021multimodal,akbari2021vatt,xu2021videoclip,zellers2021merlot} over the last few years. It has been notable that the use of pre-trained models is effective for a variety of visual downstream tasks. However, this paradigm has been adopted to a lesser extent in human motion.

We postulate that a pre-trained human motion model is not extensively used for the following reasons. First, unlike text or image data, which can be retrieved relatively readily on the web, estimating the human motion skeleton requires a significant amount of effort. While developments in pose estimation have allowed us to infer human motions from images, there still remains a hurdle in that the 3D skeleton estimated from a 2D image is not accurate, and the quality varies substantially depending on the data. On the other hand, motion capture devices can obtain highly accurate motion information, but there is a limitation in that such devices are expensive and require professional efforts for collecting a sufficient amount of clean motions. Another problem is the ambiguity of language description for motion. Human motion can semantically be interpreted in various ways. For example, ``touching a right toe", ``bending the waist", and ``touching the floor" are all valid descriptions for a single motion, let alone the polysemy in language.

%
%
\begin{figure}
\centering
\includegraphics[width=0.54\textwidth]{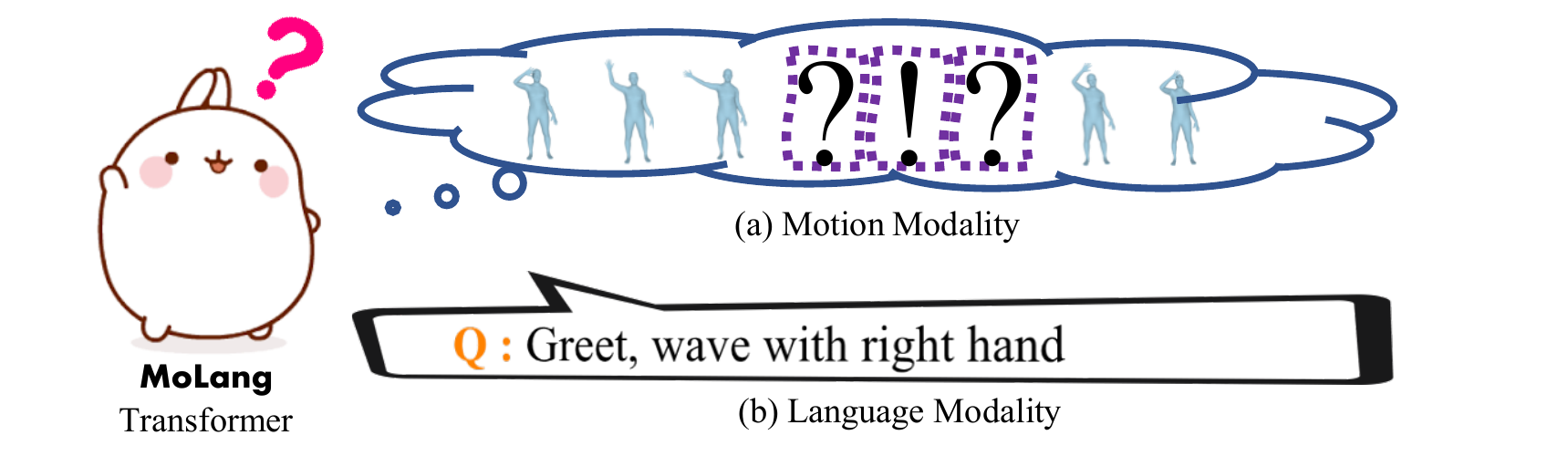}
\caption{MoLang (a Motion-Language transformer model) learns the transferable representation of human motion and language via contrastive learning with its emphasis on achieving generalizable motion representation guided by a free-form language. 
}
\label{fig:keyidea}
\end{figure}

In this work, we study these questions: Can the limited human motion data in our hands be generalized through free-formed language annotations? Can the model be adaptable with human motion and language in the wild (e.g.  YouTube~\cite{miech2019howto100m,zellers2021merlot}, Movie film~\cite{marszalek09,Sigurdsson2016HollywoodIH,Sigurdsson2017WhatAA}, DVS~\cite{lsmdc2015} etc.) to improve human motion perception in context? %
Locating and curating valid pairs of estimated human motion and free-form languages with a foundation model is an essential step forward.

As our first step, we introduce a new model MoLang that learns self-supervised human motion representations and further improves through navigating linguistic modalities. 
Enhanced with contrastive learning objective~~\cite{sohn2016improved,van2018representation} to connect motion and language, MoLang collaboratively learns human motion knowledge across modalities and benefits from free-formed human motion datasets spanning classification and description without format restrictions. 

\begin{figure*}[t!]
\centering
\includegraphics[width=0.75\textwidth]{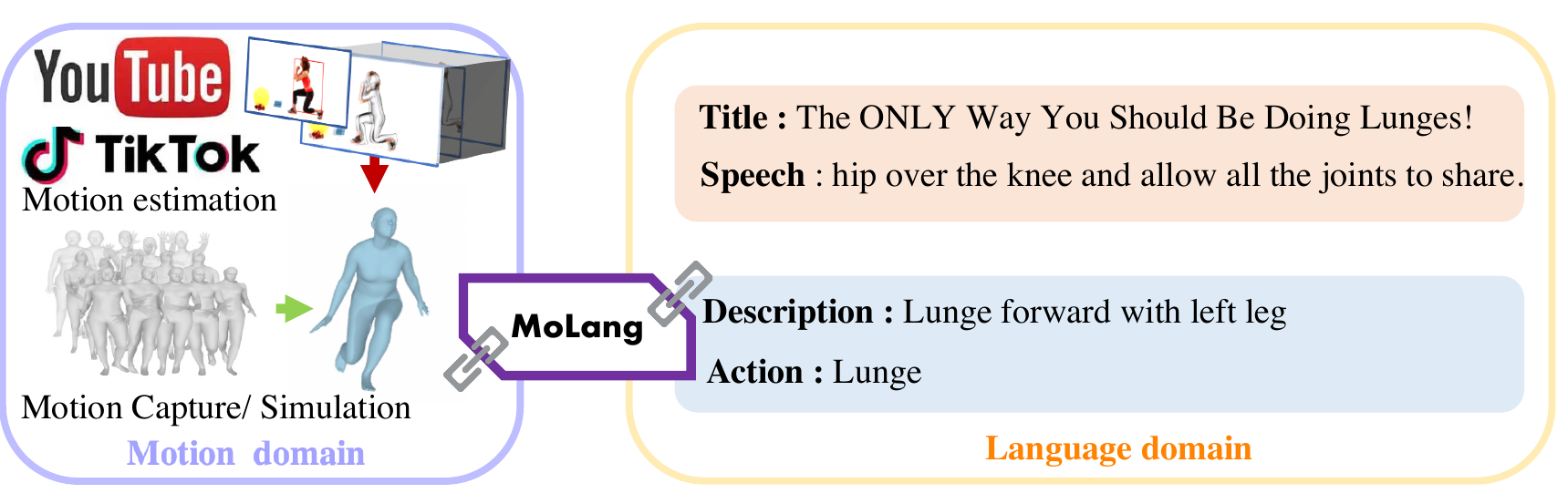}
\caption{MoLang pivots human motion and arbitrary text via solving alignment problems on both motion frames and words. Accordingly, the motion-Language co-occurrences relationships are abundantly available on the public datasets and web to potentially support training, strengthening our approach's potential directions.}
\label{fig:molang_domain}
\end{figure*}

As shown in Figure~\ref{fig:molang_domain}, MoLang seamlessly train heterogeneous motion labels (e.g.  motion description, action class) by turning them into distilled language tokens learned by language modeling objectives~\cite{devlin2018bert}. Thereby, the representation can be transferred to any human motion perception task. 
For robust motion representation, we adopted \textit{Masked Motion Prediction} and \textit{Graph Convolutional Bottleneck} to effectively align hierarchy inside a sequence of human motion structure and free-formed linguistic annotation.

The \textit{Masked Motion Prediction} is a self-supervision objective for human motion data. Considering the motion characteristics, our model learns by predicting randomly masked sub-sequence of motion without supervision. Beyond local sub-sequence prediction, we give additional hierarchy to motion encoding via \textit{Graph Convolutional Bottleneck}. We create an information bottleneck with adjacent joints to give locality by interpreting the human skeleton as a graph, which improves the sample efficiency of motion encoding. By combining these methods, we show that our model can effectively align several thousands of language phrases with motion, beyond distinguishing a few categorical labels of the human motion domain.

Experimental results show that MoLang learns robust human motion representation and significantly improves motion downstream tasks. Our model sets state-of-the-art skeleton-based action recognition tasks; HumanAct12 \cite{guo2020action2motion},
NTU-RGBD-13 \cite{guo2020action2motion}, and
UESTC \cite{ji2019large}.

Finally, ablations study and analysis of MoLang in section~\ref{sec:experiments} show that 1) multimodal alignment on motion works better when the model is trained on both motion reconstruction and contrastive cross-modal matching. 2) using a diverse set of language labels improves downstream performance compared to pre-training on the only categorical label for motion, and 3) MoLang estimates not only motion similarity but also alignment score for motion-language pair, which sets a reliable foundation for unseen human motion estimation/detection in the wild. The combination of these experimental results suggests that integrating self-supervised motion reconstruction and multimodal alignment is a promising path forward for future research on human motion understanding.

Overall, the study suggests that our newly proposed \textit{Masked Motion Prediction} objective and \textit{Graph Convolutional Bottleneck} enhances self-supervised contextualized motion representation to be more robust. Moreover, a mixture of motion-language contrastive objective with auxiliary motion reconstruction loss, which we call \textit{CstAR}, allows free-formed language annotations to guide the motion representation even more generalizable to downstream tasks.
 
Our key contributions are summarized as follows:
\begin{enumerate}
    \item We present MoLang (a Motion-Language transformer), enabling unified training of large-scale human motion datasets spanning classification and description generation as a unified resource.
    
    \item A new auxiliary reconstruction objective (\textit{Contrastive Auxiliary Reconstruction (CstAR)} for contrastive learning on motion and free-formed language, allowing our model to learn better generalizable representation. 
    
    \item A new self-supervised learning objective \textit{Masked Motion Prediction (MMP)} and an embedding module \textit{Graph Convolution Bottleneck (GCB)} for human motion modeling enable our model to navigate human joint structures efficiently.
    
    \item Experiments, ablations, and qualitative analysis demonstrate our model outperforms state-of-the-art results. We also present a novel motion retrieval benchmark to evaluate the robustness of our method.
\end{enumerate}

\section{Related Work}
\textbf{Human motion modeling.}
With the advances in motion capture device and pose estimation approaches \cite{weinzaepfel2020dope,kocabas2020vibe,rong2021frankmocap}, they enable tasks that require human skeleton data such as motion prediction or action recognition. Since human motion is spatio-temporal data constrained on the human body, it requires a model to understand both temporal and spatial features. Benefiting from the development in sequential models, there has been a significant advancement in model-based motion models. Early studies adopted Recurrent Neural Networks (RNN) to address various motion tasks \cite{fragkiadaki2015recurrent,martinez2017human,wang2019spatio,rossi2021human,zhu2016co,liu2016spatio,harvey2020robust}. In an attempt to utilize human's kinematic chain, Graph Convolutional Network (GCN) also has been employed extensively \cite{yan2018spatial,2sagcn2019cvpr,zhang2020context,cheng2020decoupling,liu2020disentangling,cheng2020skeleton,chen2021channel}.

As the Transformer \cite{vaswani2017attention} architecture has shown remarkable results in other domains, it has been introduced in the motion domain as well. Specifically, as the Transformer encoder can work as a bi-directional sequential encoder \cite{devlin2018bert}, some researches \cite{duan2021single,kim2022conditional,oreshkin2022motion} introduced Transformer-based motion generation approaches. Combined with the language command, \cite{guo2020action2motion} presented a RNN-based method to generate motion from language labels, and  \cite{petrovich2021action} used Transformer and conditional variational autoencoder (CVAE) \cite{sohn2015learning} for motion generation conditioned on language.

\textbf{Joint representation learning of motion and language.} 
Many vision and language tasks benefit from joint training on rich visual and textual data pairs. There are two main paradigms for using linguistic information to derive multimodal representation, contrastive matching, and masked prediction.
CLIP learns image classification by contrastively matching images with captions \cite{zhang2020contrastive,radford2021learning,jia2021scaling}. 
The other exploit multimodal relationships in a unified architecture.
With advent of a \emph{masked language modeling} objective \cite{devlin2018bert}, many recent works seek to uncover cross-modal interaction between objects and paired visual words in visual descriptions \cite{tan2019lxmert,li2019visualbert,lu2019vilbert,chen2019uniter,yu2020ernie}.
Recently, various pivoting relationships with language have been explored. For instance, matching video frames with visual/audio signals to transcripts \cite{rouditchenko2020avlnet,wang2021multimodal,akbari2021vatt,xu2021videoclip,zellers2021merlot} are broadly investigated, thanks to the abundance resource of web videos.

\textbf{Our work} continues these two lines of research. We adopt a joint representation training approach for human motion, which is relatively scarce. We further develop our method in a distinct way beneficial to the motion encoding domain.
Our proposed MoLang enables joint representation learning between separated domain-specific encoders.
In contrast to the recent multimodal BERT~\cite{li2019visualbert,lu2019vilbert,chen2019uniter,yu2020ernie}, we use a separate motion encoder to keep domain-specific architecture for motion, \textit{Graph Convolutional Bottleneck (GCB)} and self-supervision objective on motion, \textit{Masked Motion Prediction (MMP)} (In section~\ref{ssec:encoders}). Following this separative domain-specific design choice, we can train motion representation with MoLang with \textit{Contrastive Auxiliary Reconstruction (CstAR)} (In section~\ref{ssec:contrastiveobj}), an auxiliary motion reconstruction objective while contrastively matching sequence of words with masked motion. We further investigate the design choice in ablation study (In section~\ref{ssec:ablation}).

\begin{figure}
\centering
\includegraphics[width=1.2\linewidth]{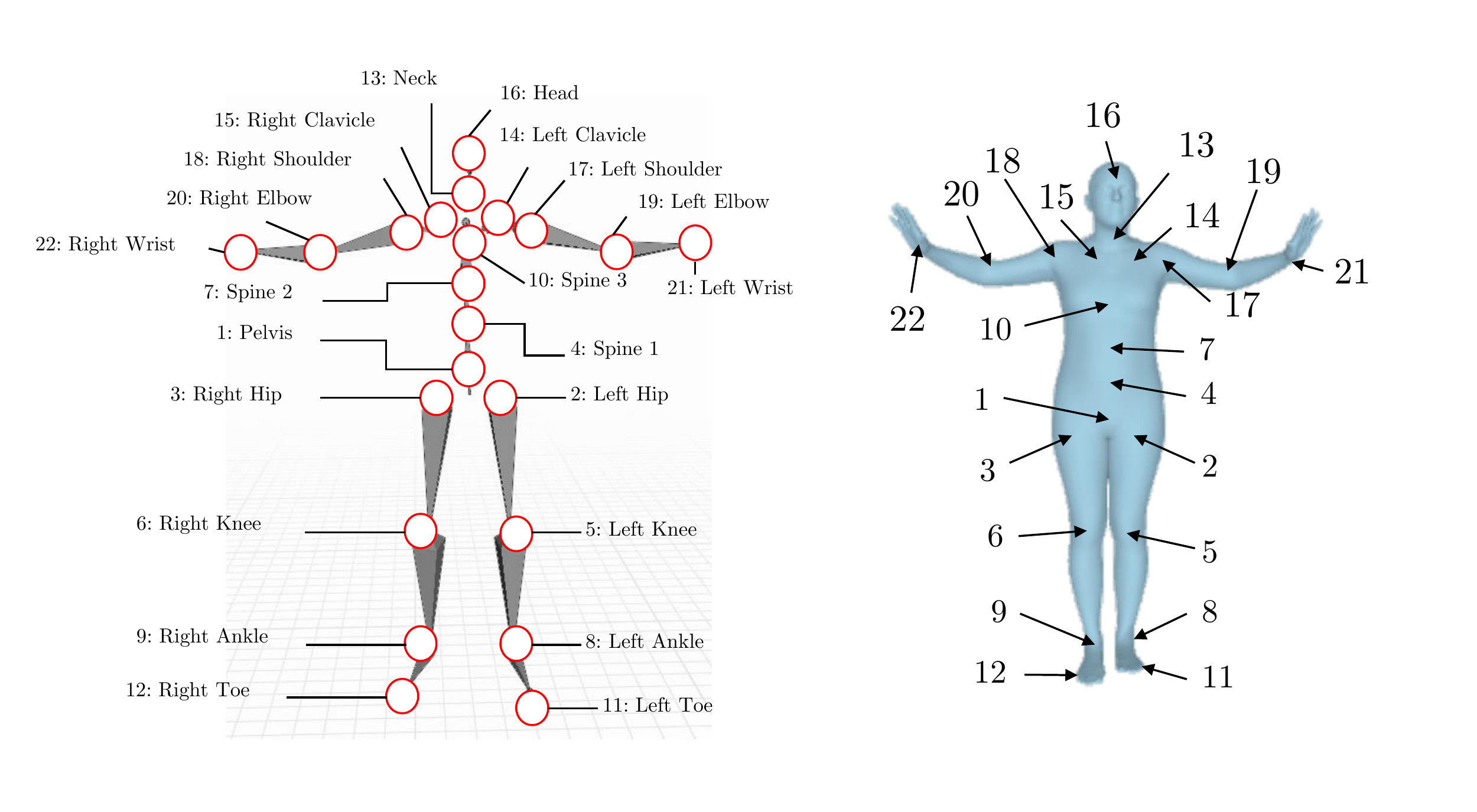}
\caption{Joint ID and Description in skeleton configuration.}
\label{fig:skeleton_id}
\end{figure}

\section{Proposed Method}
In this section, we present pre-training strategy for MoLang, including: motion representation (\ref{ssec:motion_repr}), Masked Motion Prediction (\ref{ssec:mmp}), Graph Convolutional Bottleneck (\ref{ssec:gcp}), Model architecture (\ref{ssec:modelarch}), and Contrastive pre-training objectives (\ref{ssec:contrastiveobj}).

\subsection{Motion and Language Encoding}
\label{ssec:encoders}

\subsubsection{Motion Representation}
\label{ssec:motion_repr}
We use 22 major joints to represent the human pose. Our skeleton is a subset of SMPL \cite{loper2015smpl}, which is compatible with a large-scale motion dataset \cite{AMASS:ICCV:2019}. Skeleton configuration is depicted in Figure \ref{fig:skeleton_id}.

Each joint contains relative rotation from its parent joint. Original SMPL parameterize poses with axis-angle rotation, but we adopted 6D representation \cite{zhou2019continuity} since it represents a rotation in the continuous domain, which is a preferable feature for learning in neural networks. For motion length with $T$, whole sequence can be written in $M \in \mathbb{R}^{T \times 22 \times 6 }$. We downsample motion if the frames per second (fps) of motion is greater than 30Hz. If fps is less than 30 fps, motion is upsampled using interpolation in 6D representation so that the model can consistently observe 30 fps of motion data. For the application in the downstream task, we prepend \textit{\textlangle CLS\textrangle} token to be the starting token of motion. Therefore, input motion is represented as $M_{\textrm{input}} \in \mathbb{R}^{(T+1) \times 22 \times 6}$.

\subsubsection{Masked Motion Prediction}
\label{ssec:mmp}
On the basis of Transformer \cite{vaswani2017attention} encoder architecture, we train motion encoder with masked prediction strategy \cite{devlin2018bert}. Since human motion is highly correlated with nearby frames, we mask a series of frames instead of masking single frame in motion. We first sample the masking motion length $l$ from uniform distribution $l \sim \mathcal{U}(1, 30)$ and set the start frame of masking sequence $t_{\textrm{start}}$ from $t_{\textrm{start}} \sim \mathcal{U}(0, T-l)$ where $T$ is the maximum length of motion.
Then, masking frames $[t_{\textrm{start}}, t_{\textrm{start}+l}]$ will be replaced by Gaussian noise to mask frames. Since motion input is intrinsically represented in the continuous space, we do not use separate tokenization in the motion processing pipeline. When given motion is shorter than maximum motion length $T$, we pad remaining frames with zero and introduce \textit{valid segmentation embedding} to make the model distinguish valid frames from padded one. This is a simple embedding of valid frames marked with one and padded frames marked with zero. In the end, input motion, positional embedding, and valid segmentation embedding are summed as a Transformer's input. 
We train this \textit{Masked Motion Prediction (MMP)} objective on BERT for human motion encoding and also refer to it as MMP-BERT.

\begin{figure}
\centering
\includegraphics[width=0.5\textwidth]{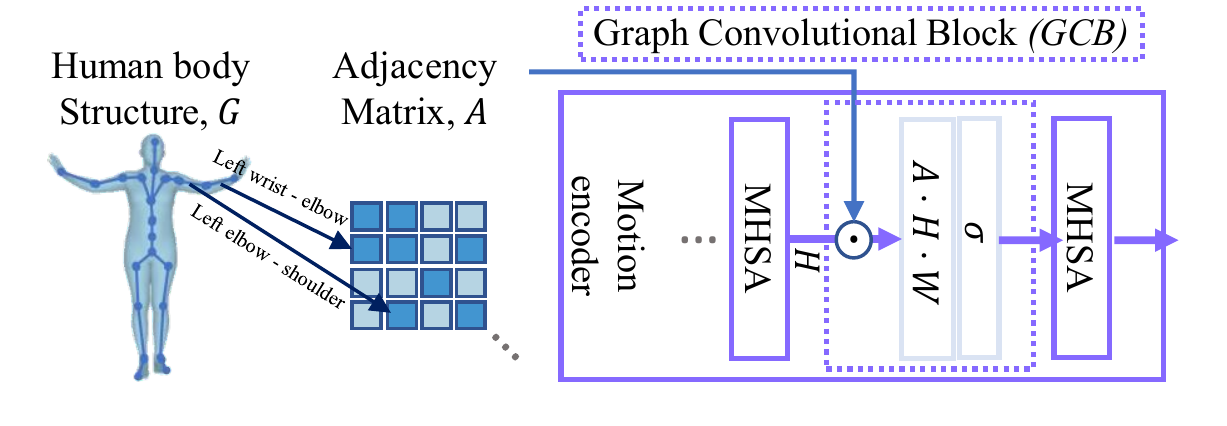}
\caption{Adapting Human body skeleton $G=\{V:Joints,E:Bones\}$ for human motion sequence encoding.
We put a \textit{Graph Convolutional Bottleneck, (GCB)} between Multi-Head Self-Attention (MHSA) Blocks~\cite{vaswani2017attention} in our motion encoder, MMP-BERT. }
\label{fig:motion_encoder}
\end{figure}

\subsubsection{Graph Convolutional Bottleneck}
\label{ssec:gcp}

The human body can be interpreted as a graph structure. We can consider (Joints, Bones) in the human body as $G = \{\textrm{Nodes}, \textrm{Edges}\}$. 
Node is denoted as $\mathcal{J} = \{ j_1, j_2, \cdots, j_J \}$, where $J$ is the number of joints. Adjacency matrix $\mathbf{A} = (a_{ik}) \in \mathbb{R}^{J \times J}$ represent correlation between joints $j_{i}$ and $j_{k}$. Neighboring joints of $j_i$ is denoted as:
\begin{equation}
\mathcal{N}(j_{i}) = \left\{ j_{k} \mid a_{ik} \neq 0 \right\}  
\end{equation}
We define the elements of the adjacency matrix to have one if joints are connected and zero otherwise. For the feature dimension of $d$, the output of  $L$-th encoder layer $\mathbf{H}^{L} \in \mathbb{R}^{J \times d}$ is computed in the graph convolutional layer as:

\begin{equation}
    \mathbf{y}_{i} = \sigma \left( \sum_{j_k \in \mathcal{N}(j_i)} a_{ik} \mathbf{h}^{L}_{k} \mathbf{W} \right)
\end{equation}
where $\mathbf{W} \in \mathbb{R}^{d \times d}$ is a trainable weight and $\sigma$ is a non-linear activation function. We insert the graph convolutional layer between Multi-Head Self-Attention Blocks~\cite{vaswani2017attention} to prioritize relation in the group of neighboring joints. 
Our motion encoder architecture is shown in Figure \ref{fig:motion_encoder}.

\subsubsection{Language Encoding}
We use BERT \cite{devlin2018bert} architecture as our text encoder. Instead of training from scratch, we initialize the text encoder with `bert-base-uncased' pre-trained weights provided from HuggingFace Transformers~\cite{wolf-etal-2020-transformers}. Language transformer is also finetuned in our contrastive learning pipeline.

\subsection{Joint Model Architecture}
\label{ssec:modelarch}

\begin{figure*}[t]
\centering
\includegraphics[width=0.80\textwidth]{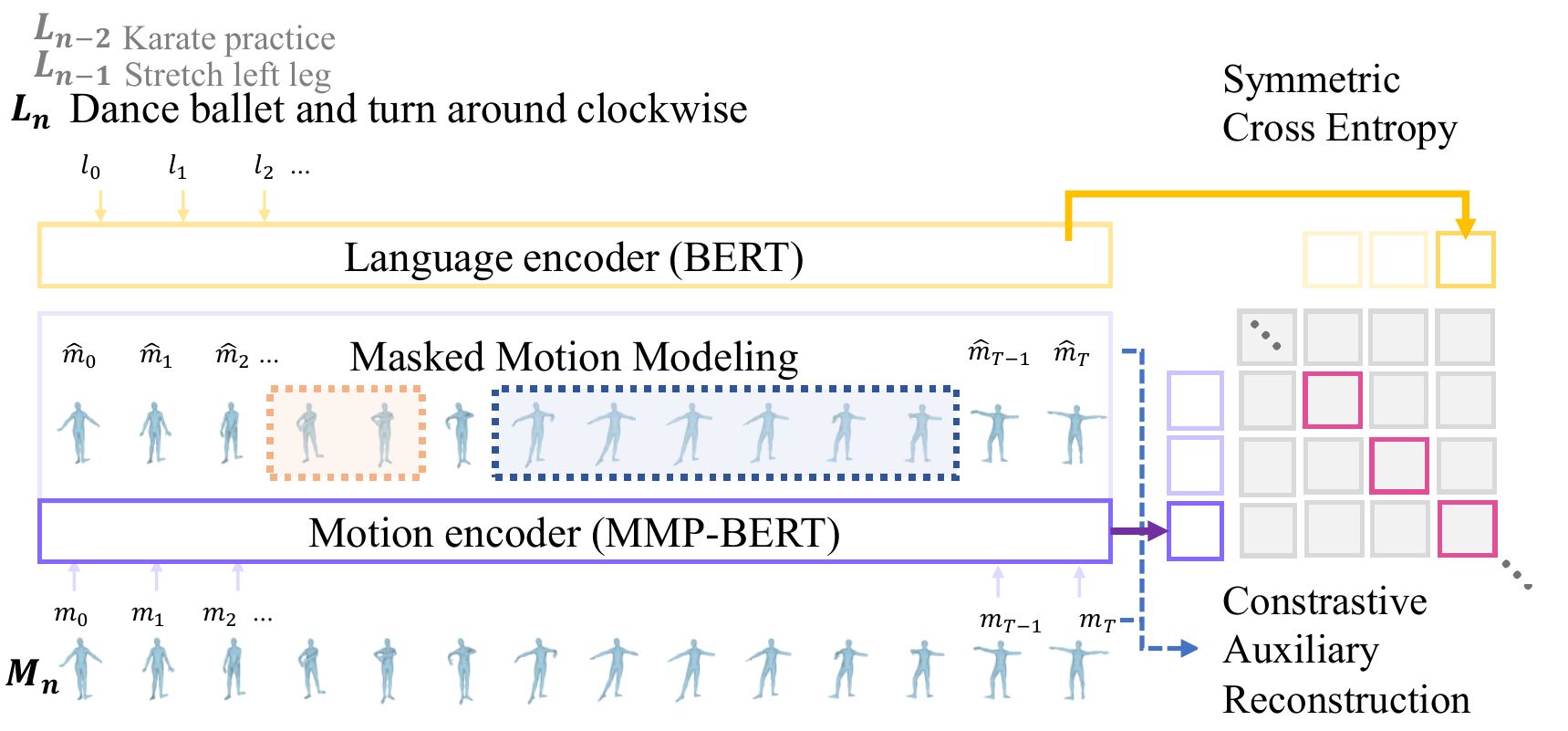}
\caption{An overview of MoLang training pipeline. We provide sequence-level representations of human motion to a Motion encoder. The motion encoder contextualizes over fine-grained motion sequence with Masked Motion Prediction objective and further guided with Symmetric Cross-Entropy by pre-trained Language encoder. The model must maximize its similarity of language label and motion encoding while keeping it contextualized.}
\label{fig:molang}
\end{figure*}

An overview of MoLang is shown in Figure~\ref{fig:molang}. MoLang is trained on the top of the pre-trained motion encoder (MMP-BERT) discussed in \ref{ssec:encoders}. Each motion and language modality is first encoded separately by a modality-specific pre-trained transformer encoder. Then, the representations will be trained with contrastive loss. Our model use transformer architecture as motion encoder and contrastive loss with \textit{Auxiliary Reconstruction loss}.

\subsubsection{Contrastive learning with Auxiliary Reconstruction (CstAR)}
\label{ssec:contrastiveobj}

We use contrastive learning with the InfoNCE loss \cite{sohn2016improved,van2018representation} with auxiliary motion reconstruction loss $\mathcal{L}_{\textrm{recon}}$. Auxiliary motion reconstruction loss is an L1-norm between the output of the Motion encoder and input motion. The auxiliary term makes our motion encoder predict masked motion while adapting cross-modal matching (i.e., contrastive learning) between motion and language domain. 
We optimize the \textit{CstAR} loss:
{\small
\begin{align}\label{eq:cst}
&\mathcal{L}_{CstAR}(M, L)  = \nonumber \\
&\!\!\!\sum_{i} \left(
\frac{\exp{ \left( s(\bm{m}^{(i)}, \bm{l}^{(i)}) / \tau \right)}}{ \sum_{\bm{m}}\exp{ \left( s(\bm{m}, \bm{l}^{(i)}) / \tau \right)}} +
\frac{\exp{ \left( s(\bm{m}^{(i)}, \bm{l}^{(i)}) / \tau \right) }}{ \sum_{\bm{l}}\exp{ \left( s(\bm{m}^{(i)}, \bm{l}) /\tau \right)}} + \alpha \mathcal{L}_{\textrm{recon}} \right) 
\nonumber \\
&\mbox{where} \;\; \mathcal{L}_{\textrm{recon}} = \left\lVert \bm{m}^{(i)} - \bm{m}^{(i)}_{\textrm{input}} \right\rVert_1
\end{align}
}%
where $\tau$ is a learnable temperature, $\alpha$ is a weight for auxiliary motion reconstruction loss, and $\bm{m}^{(i)}, \bm{l}^{(i)}$ are vector representations of the co-occuring motion-language pair $(m^{(i)}, l^{(i)})$ which are encoded by $g_M(m^{(i)})$ and Language encoder $g_L(l^{(i)})$, Motion encoder respectively; $s( *, *)$ computes the scaled cosine similarity. 
Vector similarities across modalities are optimized to reconstruct co-occurrence patterns in training corpora.

\section{Experiments}
\label{sec:experiments}
This section describes the datasets and experiment settings and demonstrates the experimental results of our model.

\subsection{Datasets}
\label{subsec:dataset} 
The following datasets are used in our experiment. All datasets used in our experiments follows SMPL \cite{loper2015smpl} representation format. Unlike the 3D poses from AMASS \cite{AMASS:ICCV:2019} and BABEL \cite{BABEL:CVPR:2021}, which are obtained from accurate motion capture equipment, 3D pose of other datasets are obtained by pose estimation framework such as SMPLify \cite{Bogo:ECCV:2016} or VIBE \cite{kocabas2020vibe}.

\textbf{AMASS \cite{AMASS:ICCV:2019}} This is a large collection of motion dataset. Datasets collected in this dataset are obtained from a highly accurate optical marker-based motion capture system. It integrates more than 50-hours mocap datasets by using widely used SMPL format.

\textbf{BABEL \cite{BABEL:CVPR:2021}} This is a language annotation for the AMASS dataset. Unlike most previous motion labels, which only describe motions with high-level semantic labels, this provides frame-level annotation. Also, it describes motion closer to natural language rather than using single-word categorical labels.

\textbf{HumanAct12 \cite{guo2020action2motion}}
HumanAct12 dataset is a annotated version of PHSPD \cite{zou20203d} dataset with temporal cropping. In order to obatin 3D poses, it employed multi-step fitting by OpenPose \cite{cao2017realtime} and SMPLify \cite{Bogo:ECCV:2016}. It comprises 1,191 motions with 12 labels.

\textbf{Refined NTU-RGBD 13 \cite{guo2020action2motion}}
This dataset is a subset of NTU-RGBD \cite{shahroudy2016ntu} action recognition dataset, published by \cite{guo2020action2motion}. It is a re-estimationion of 3D poses through VIBE \cite{kocabas2020vibe} to suppress noise in original NTU-RGBD dataset. This comprises 3,902 motion clips with 13 action classes.

\textbf{UESTC \cite{ji2019large}} This dataset consists of 40 aerobic exercise categories. VIBE \cite{kocabas2020vibe} is employed to estimate SMPL parameters from the original dataset. We split train and test by cross-subject recognition protocol, motions recorded from 51 subjects for the train split and the others for the test split.

\subsection{Implementation Details}
\subsubsection{Preprocessing}
Since motion datasets have various frame rates, we make every motion have the same (30 fps) frame rates. Interpolation is conducted in 6D continuous representation \cite{zhou2019continuity}. The maximum length of motion is set to 150 frames (5 seconds). The preprocessed 364K samples in AMASS \cite{AMASS:ICCV:2019} are used to pre-train the motion encoder.

In the BABEL dataset, we use frame-level processed annotation for contrastive learning. We exclude motions annotated as `transition' since it does not convey consistent semantic information for describing motion. Unique language descriptions used in the training set are about five thousand.

\subsubsection{Hyperparameters}
Our motion encoder is a ten-layer transformer encoder with 12 heads and 1,024 feedforward dimensions. Graph convolutional bottleneck is located between the fourth and fifth transformer encoder layers. The maximum context length is set to 150. In the contrastive training setting, the output of the motion encoder is projected to 768 dimensions to match with the text encoder. To process text inputs, we use pre-trained BERT \cite{devlin2018bert}. Learnable temperature is used to compute symmetric cross-entropy (SCE), initialized with 0.07. Weights for $\mathcal{L}_{\textrm{recon}}$ is set to 10.0.

\subsubsection{Training}

Our model takes two training stages. First, our model is trained in a self-supervised manner with motion capture datasets (i.e.  AMASS~\cite{AMASS:ICCV:2019}). In the second stage, MoLang is trained on BABEL motion-language paired training dataset together using \textit{CstAR}.

\subsection{Sekelton-based Action Recognition}

To validate the effectiveness of our motion representation, we evaluate our model on skeleton-based action recognition downstream tasks. As our text encoder can process natural languages, we directly input the class label of the target dataset to our text encoder, which is an advantage over other classifiers that are fixed to class labels at the time of training. In this transfer learning setting, we finetune on target dataset with the same contrastive objective. We evaluate our approach on the HumanAct12, Refined NTURGBD13, and UESTC in Table \ref{table:action-recognition}. 

\begin{table*}
\centering
\begin{tabularx}{14.7cm}{ 
    >{\raggedright\arraybackslash}m{6cm}
  | >{\centering\arraybackslash}m{2.3cm} 
  | >{\centering\arraybackslash}m{2.3cm} 
  | >{\centering\arraybackslash}m{2.3cm}}
 \hline
 Methods & HumanAct12 & NTU-RGBD13 & UESTC \\
 \toprule \hline
 LSTM & 68.07 & 80.58 & 78.33 \\ \hline
 P-LSTM \cite{shahroudy2016ntu} & 70.20 & 82.03 & 80.14 \\ \hline
 TCN \cite{lea2017temporal} & 75.21 & 84.54 & 85.77 \\ \hline
 2s-AGCN \cite{2sagcn2019cvpr} &79.01 & 90.63 & 90.86 \\ \hline
 2s-AGCN \cite{2sagcn2019cvpr} \scriptsize{(pretrain)} & 83.19 & 91.67 & 95.63 \\ \hline
 DC-GCN \cite{cheng2020decoupling} & 82.72 & 90.75 & 90.51 \\ \hline
 DC-GCN \cite{cheng2020decoupling} \scriptsize{(pretrain)} & 85.46 & 92.71 & 96.58 \\  \specialrule{1.2pt}{0pt}{0pt}
 MMP-BERT & 84.87 & 85.77 & 90.76 \\ \hline
 MMP-BERT \scriptsize{(pretrain)} & 88.82 & 91.85 & 94.74 \\ \hline
 MoLang & 91.48 & 93.86 & 96.97 \\ \hline
\end{tabularx}
\caption{Comparison of accuracy on skeleton-based action recognition task. Pre-trained models are trained on BABEL \cite{BABEL:CVPR:2021} dataset.}
\label{table:action-recognition}
\end{table*}

\begin{figure}[t]
\centering
\includegraphics[width=0.5\textwidth]{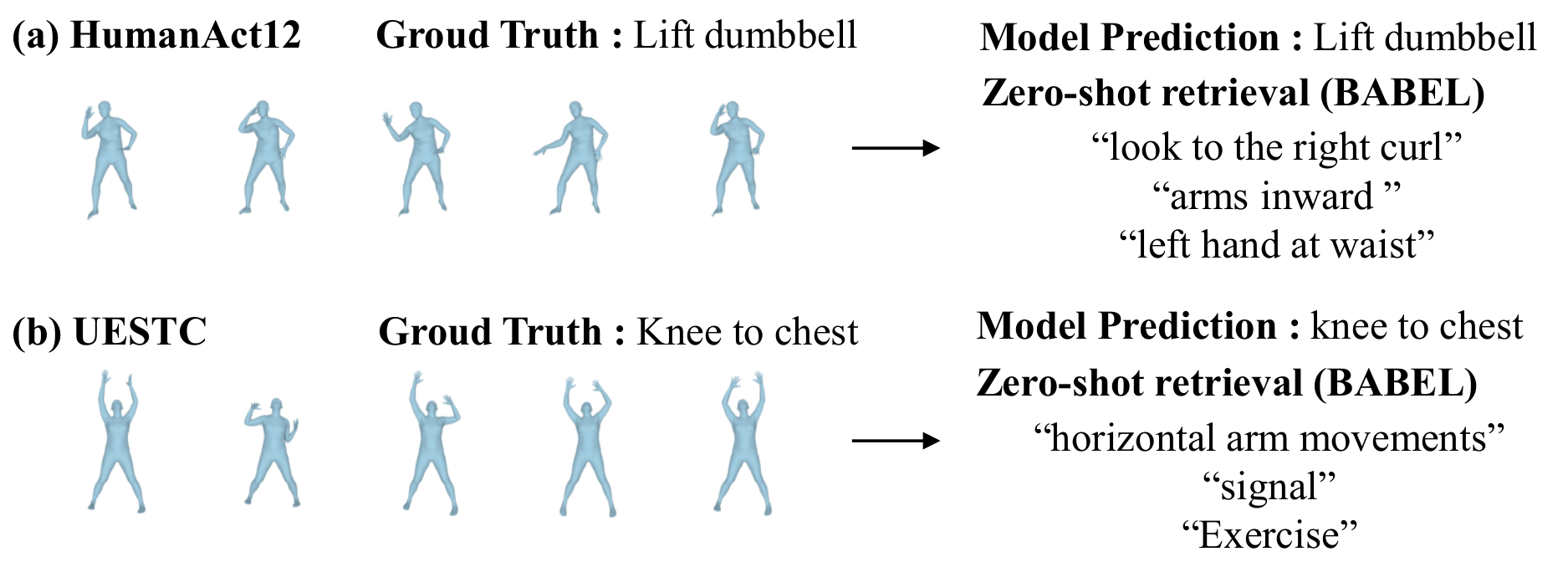}
\caption{Qualitative results on action recognition downstream tasks, HumanAct12 and UESTC.}
\label{fig:qual-act-downstream}
\end{figure}

Since GCN-based action recognition models take $xyz$-coordinates in Euclidean space, we convert our 6D representation to joint positions using \cite{loper2015smpl} to make a fair comparison for the baseline models. Both Deep LSTM and P-LStM \cite{shahroudy2016ntu} is based on LSTM architecture and used to evaluate action recognition in large-scale RGB+D action recognition dataset \cite{shahroudy2016ntu}. Deep LSTM is traditional RNN-based architecture, and P-LSTM is a part-aware LSTM model, which splits the human body's motion into predefined parts. TCN \cite{lea2017temporal} is a multi-layer 1-dimensional convolution network, which takes convolution operation on the temporal dimension. The 2s-AGCN \cite{2sagcn2019cvpr} is a graph convolution-based architecture that conducts convolutions on both spatial and temporal dimensions with two streams of networks. Decouple GCN (DC-GCN) \cite{cheng2020decoupling} break channels into several groups to enhance spatial aggregation. MMP-BERT is a Transformer encoder layer trained with our masked motion prediction strategy. The pre-trained version is trained on BABEL \cite{BABEL:CVPR:2021} dataset before downstream tasks.

\begin{figure}[t]
\centering
\includegraphics[width=0.51\textwidth]{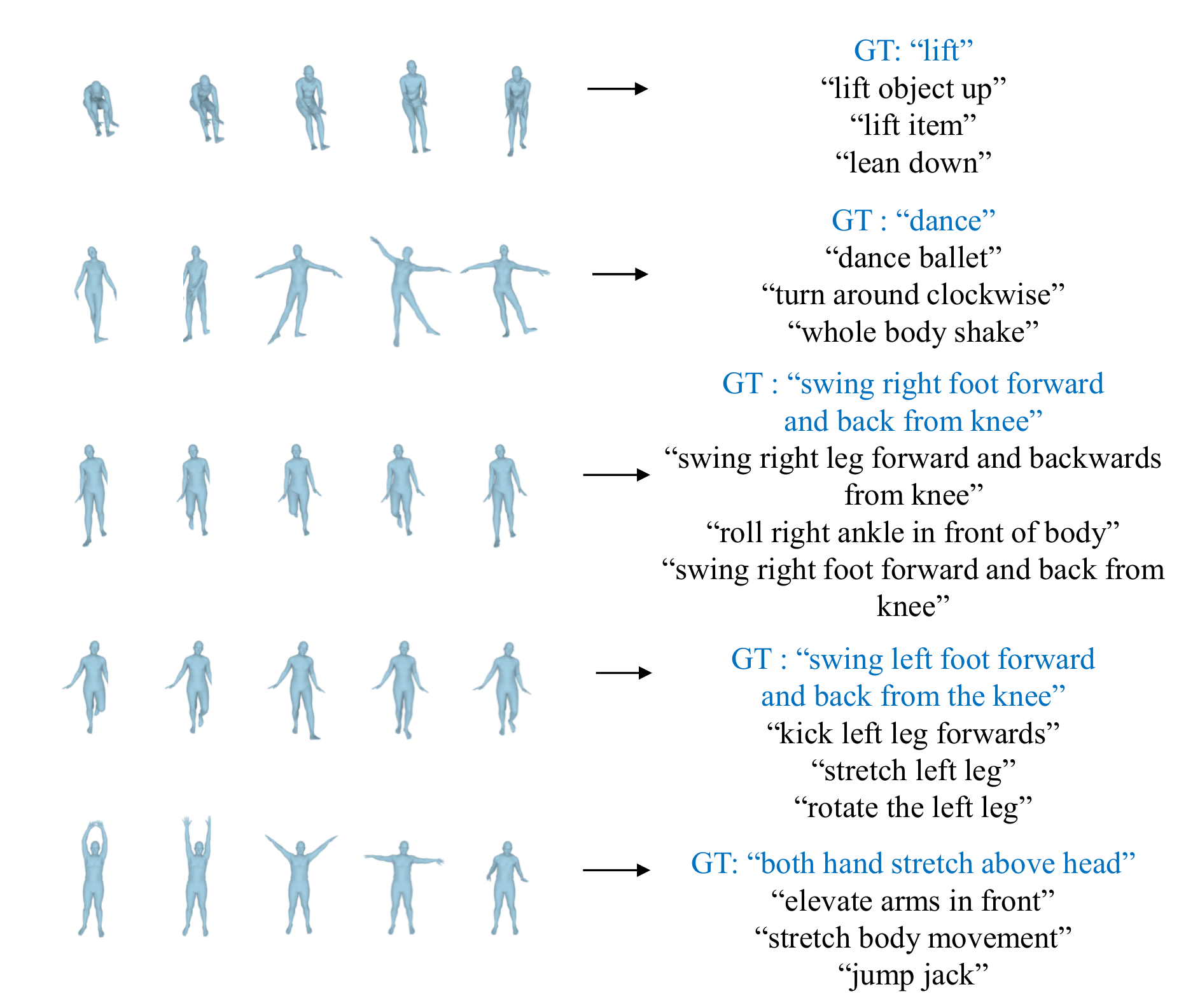}
\caption{Qualitative results on motion representation. The closest three language descriptions are chosen from BABEL datset.}
\label{fig:qual-act-reco}
\end{figure}

To evaluate the capability of language description, we also test our model on a separate set of BABEL for qualitative evaluation. We give more than five thousand labels that depict motion to the text encoder and compare the labels with the top 3 similarities with motion. This is shown in Figure \ref{fig:qual-act-reco}.

\begin{table}
\centering
\begin{tabularx}{8.56cm}{ 
    >{\centering\arraybackslash}m{0.7cm}
  | >{\centering\arraybackslash}m{0.7cm} 
  | >{\centering\arraybackslash}m{0.76cm} 
  | >{\centering\arraybackslash}m{1.3cm}
  | >{\centering\arraybackslash}m{1.3cm}
  | >{\centering\arraybackslash}m{1.3cm}}
 \hline
 \tiny{$MMP$} & \tiny{$GCB$} & \tiny{$CstAR$} & Human\ Act12 & NTU-RGBD13 & UESTC \\
 \toprule \hline
 $-$ & $-$ & $-$ & 82.35 & 88.02 & 91.79 \\
 \hline
 $\checkmark$ & $-$ & $-$ & 90.31 & 92.51 & 95.26 \\ \hline
 $-$ & $\checkmark$ & $-$ & 84.2 & 88.84 & 91.93 \\ \hline
 $-$ & $-$ & $\checkmark$ & 82.65 & 88.75 & 92.73 \\ \hline
 $\checkmark$ & $\checkmark$ & $-$ & 90.33 & 92.53 & 95.39 \\ \hline
 $\checkmark$ & $-$ & $\checkmark$ & 91.07 & 92.67 & 95.37 \\ \hline
 $\checkmark$ & $\checkmark$ & $\checkmark$ & 91.48 & 93.86 & 96.97 \\
\hline
\end{tabularx}
\caption{(Action recognition) Ablation study on self-supervised pre-training of Masked Motion Prediction (MMP), Graph Convolutional Bottleneck (GCB), and Auxiliary Reconstruction loss (CstAR).}
\label{table:ac-recog-ablation}
\end{table}

\subsection{Motion Retrieval}

\begin{figure}[t]
\centering
\includegraphics[width=0.5\textwidth]{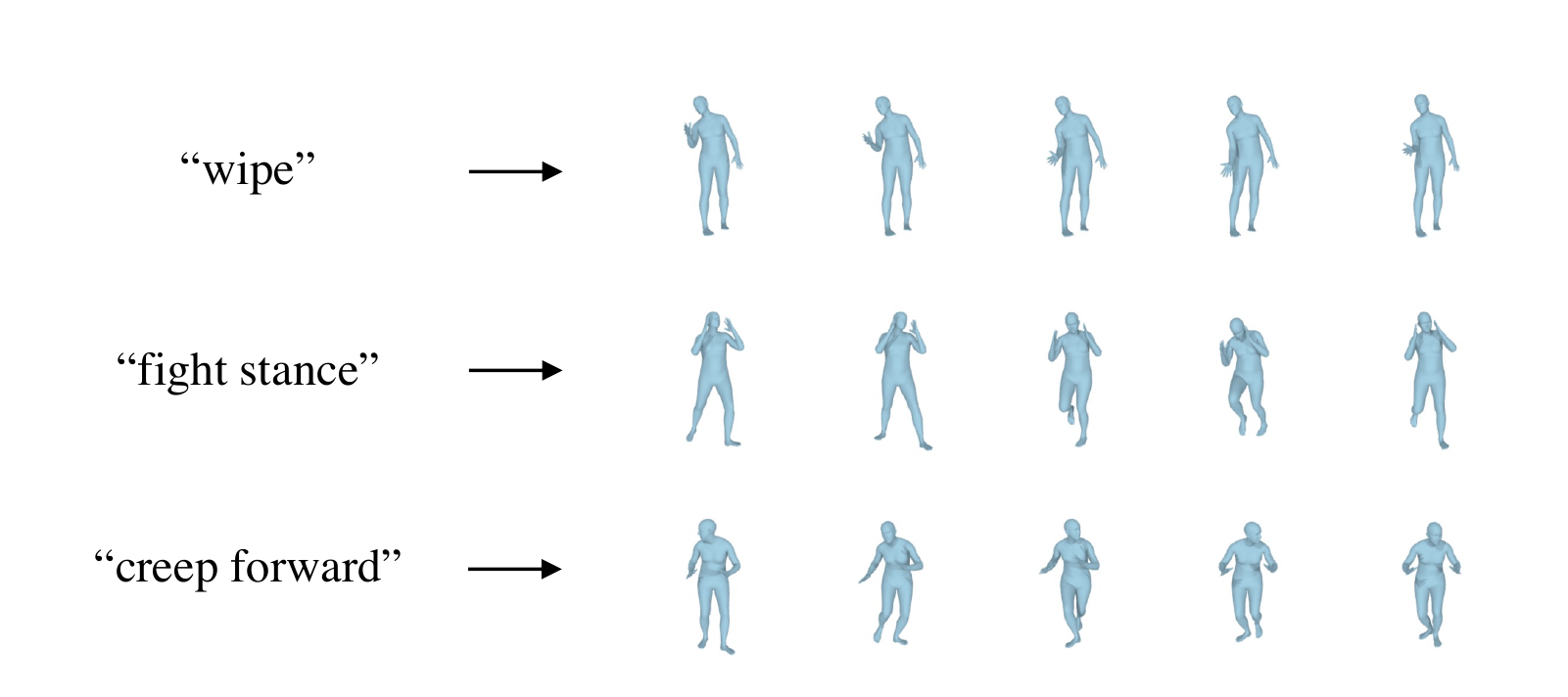}
\caption{Qualitative results on motion retrieval. The closest motion for given label is retrieved.}
\label{fig:qual-mr}
\end{figure}

We make multiple choice motion retrieval questions from the BABEL dataset for the motion retrieval task. First, we choose frequent and visually exclusive 15 labels (e.g., sit, dance, wipe, crouch, etc.) from the entire labels. Then, motion is categorized by those labels to compose candidates for each label. At the evaluation phase, the model is asked to find the closest motion for the given text among 15 randomly chosen motions. We compose 450 motion retrieval questions to evaluate motion retrieval performance. Qualitative results are presented in Figure~\ref{fig:qual-mr}.

\subsection{Ablation study}
\label{ssec:ablation}
We also conduct an ablation study to empirically validate our self-supervised pre-training design decisions on motion, Graph Convolutional Bottleneck, and Auxiliary Reconstruction loss for contrastive learning (CstAR). 

Results are in Table~\ref{table:ac-recog-ablation} (Sekelton-based Action Recognition) and Table~\ref{table:motion-retrieval-ab} (Motion Retrieval)

\textbf{Motion-only self-supervised pre-training helps contrastive learning.}
We pre-train our motion encoder in a self-supervised fashion before contrastive pre-training, and we call this \textit{MMP} pre-training here. (details in section~\ref{ssec:encoders})
MMP pre-training always helps in downstream and retrieval when preceded beforehand. This confirms pre-training power of domain encoder has an essential role in robust contrastive learning.

\textbf{\textit{GCB} helps motion representation.}
We employ an adjacency matrix-based graph convolutional layer to make an information bottleneck, which encourages the model to focus on neighboring joints over distant ones. Specifically, combined with \textit{CstAR}, \textit{GCB} significantly improves model performance. (6th row \textit{vs} 7th row) We assume that the model learns meaningful priors for the understanding sequence of human poses from using \textit{GCB}.

\textbf{\textit{CstAR}, Auxiliary reconstruction loss helps robust cross-modal matching.}
Comparing (\textit{MMP} + \textit{GCB}, 5th row) with full model (7th row) in Table~\ref{table:ac-recog-ablation},\ref{table:motion-retrieval-ab}, we see the accuracy increases in both classification and retrieval. Even simple contrastive learning (1st row) improves with the $CstAR$ objective (4th row). We believe two advantages can explain the improvements. One reason is,  The loss helps the model to pay attention on motion completion when updating for contrastive learning.
The other reason stems from the motion augmentation effect during contrastive learning. This also confirms that the auxiliary reconstruction loss for masked motion guides robust contrastive matching during motion-language alignment. 


\begin{table}
\centering
\begin{tabularx}{8.2cm}{ 
    >{\centering\arraybackslash}m{0.8cm}
  | >{\centering\arraybackslash}m{0.8cm} 
  | >{\centering\arraybackslash}m{0.8cm} 
  | >{\centering\arraybackslash}m{1.8cm}
  | >{\centering\arraybackslash}m{1.8cm}}
 \hline
 \tiny{$MMP$} & \tiny{$GCB$} & \tiny{$CstAR$} & Top-1 (\%) & Top-3 (\%)\\
 \toprule \hline
 $-$ & $-$ & $-$ & 48.44 & 74.22\\\hline
 $\checkmark$ & $-$ & $-$ & 55.18 & 75.14 \\ \hline
 $-$ & $\checkmark$ & $-$ & 49.64 & 74.18 \\ \hline
 $-$ & $-$ & $\checkmark$ & 50.24 & 74.39\\ \hline
 $\checkmark$ & $\checkmark$ & $-$ & 55.06 & 75.25 \\ \hline
 $\checkmark$ & $-$ & $\checkmark$ & 56.12 & 76.62 \\ \hline
 $\checkmark$ & $\checkmark$ & $\checkmark$ & 56.76 & 77.34 \\
\hline
\end{tabularx}
\caption{Ablation study on self-supervised pre-training of Masked Motion Prediction (MMP), Graph Convolutional Bottleneck (GCB), and Auxiliary Reconstruction loss (CstAR). Top-1 and Top-3 accuracy on motion retrieval task.}
\label{table:motion-retrieval-ab}
\end{table}

\section{Conclusion}
We introduce MoLang, which learns self-supervised representations jointly through motion and language modalities. We have found empirically that our model achieves robust representation in motion-language matching. Furthermore, we strengthen our contrastive matching objective by utilizing auxiliary reconstruction loss. Our model performs well in both downstream action recognition benchmarks and motion-text retrieval settings.


Moreover, the motion-language similarity evaluation metric, available from our MoLang, is an essential foundation for leveraging human motion-language pairs in the wild, such as web videos and movie films. Specifically, curating and incorporating web-scale multimodal data (e.g., pairs of estimated human motion and noisy subtitles) for effective human motion understanding is a promising direction and one that has significant space for future work. 

\bibliography{aaai23}

\clearpage

\appendix

\section{Appendix}
We present the following items in the supplemental:
\begin{enumerate}
    \item More Qualitative Results
    \item t-SNE Visualization of Language Representations
    \item Hyperparameters
\end{enumerate}

\section{Appendix 1: More Qualitative Results}
\label{supp:qualitativeanalysis}

\subsection{More Skeleton-based Action Recognition Results}
\label{supp:moreaction}
\begin{figure*}[ht]
\centering
\includegraphics[width=\textwidth]{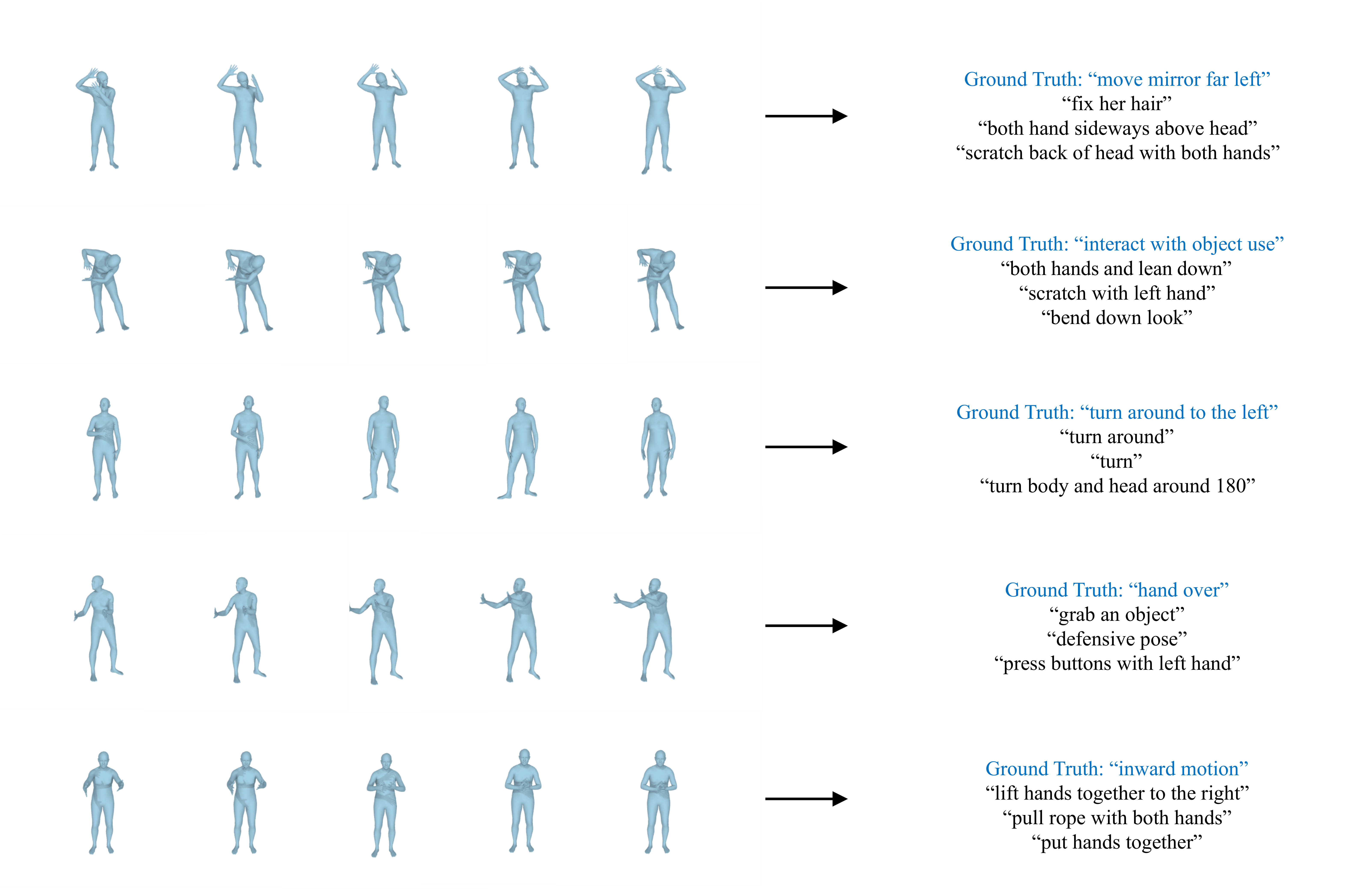}
\vspace{10mm}
\caption{Action recognition for BABEL dataset. Top-3 predicted labels from BABEL dataset.}
\label{fig:app_action_recognition_babel}
\end{figure*}

\begin{figure*}[ht]
\centering
\includegraphics[width=\textwidth]{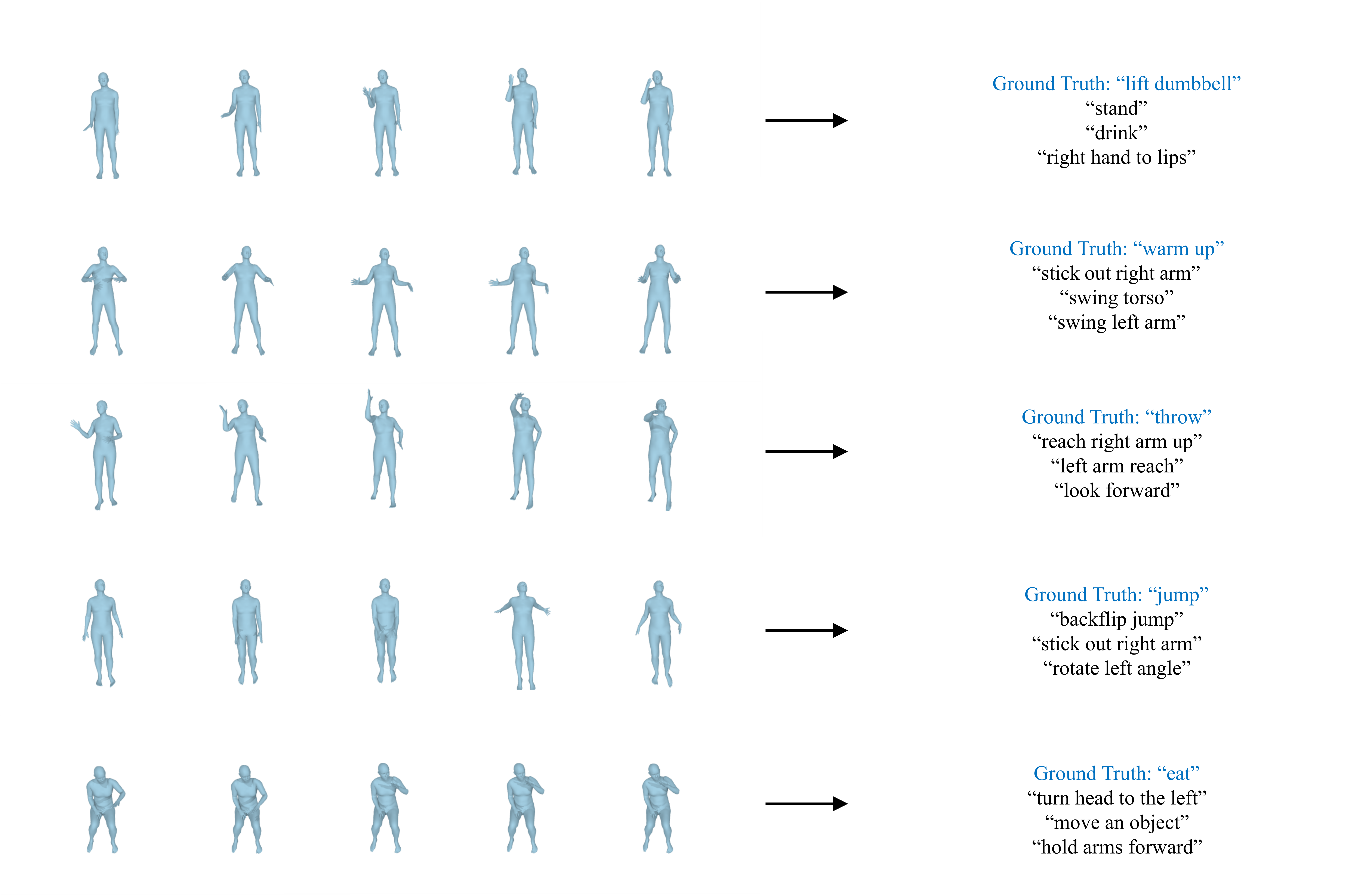}

\caption{Action recognition for HumanAct12 dataset. Top-3 predicted labels from BABEL dataset.}
\label{fig:app_action_recognition_humanact12}
\end{figure*}

\begin{figure*}[!ht]
\centering
\includegraphics[width=\textwidth]{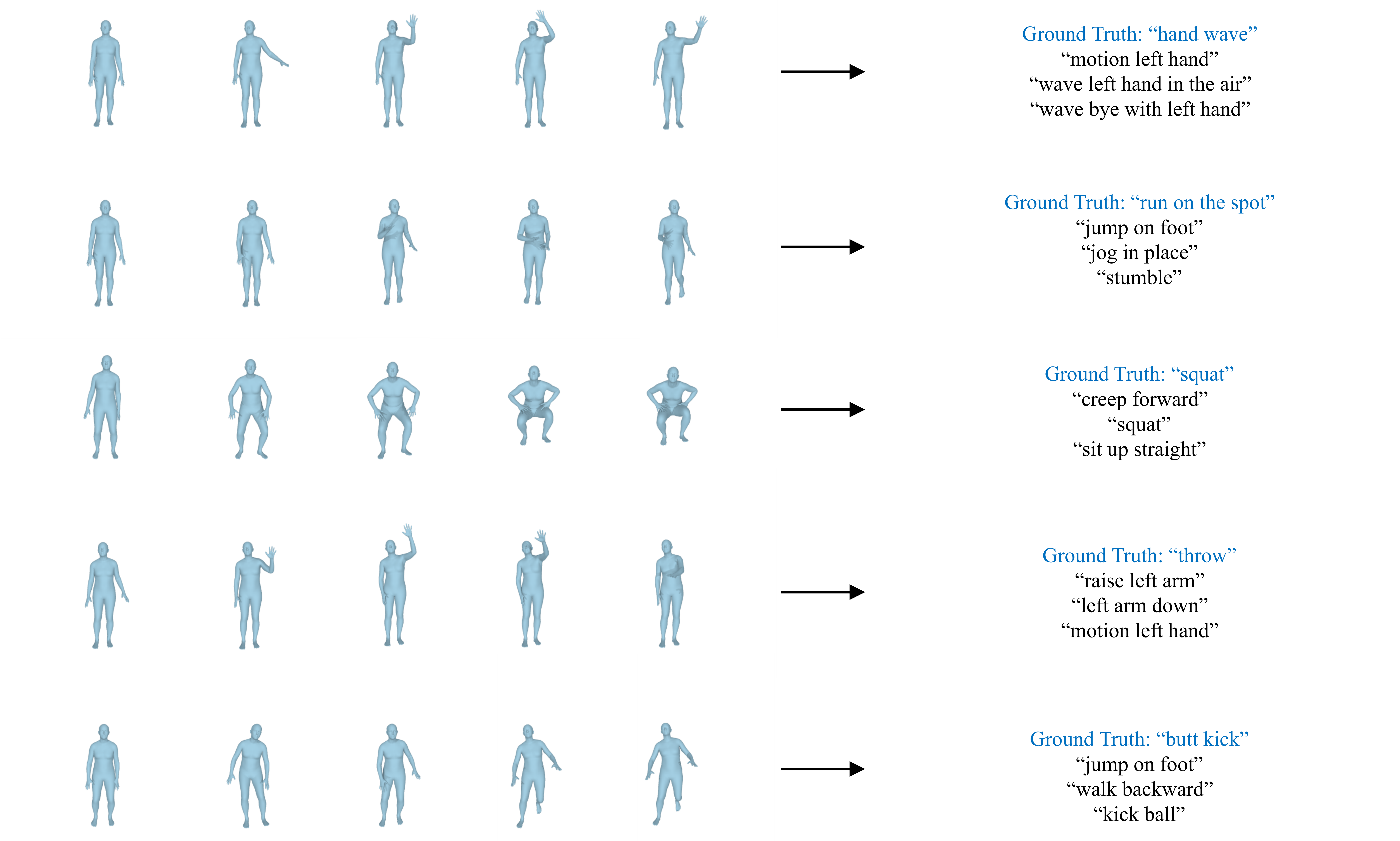}
\caption{Action recognition for NTU-RGBD 13 dataset. Top-3 predicted labels from BABEL dataset.}
\label{fig:app_action_recognition_nturgbd13}
\end{figure*}

\begin{figure*}[!ht]
\vspace{-10mm}
\centering
\includegraphics[width=\textwidth]{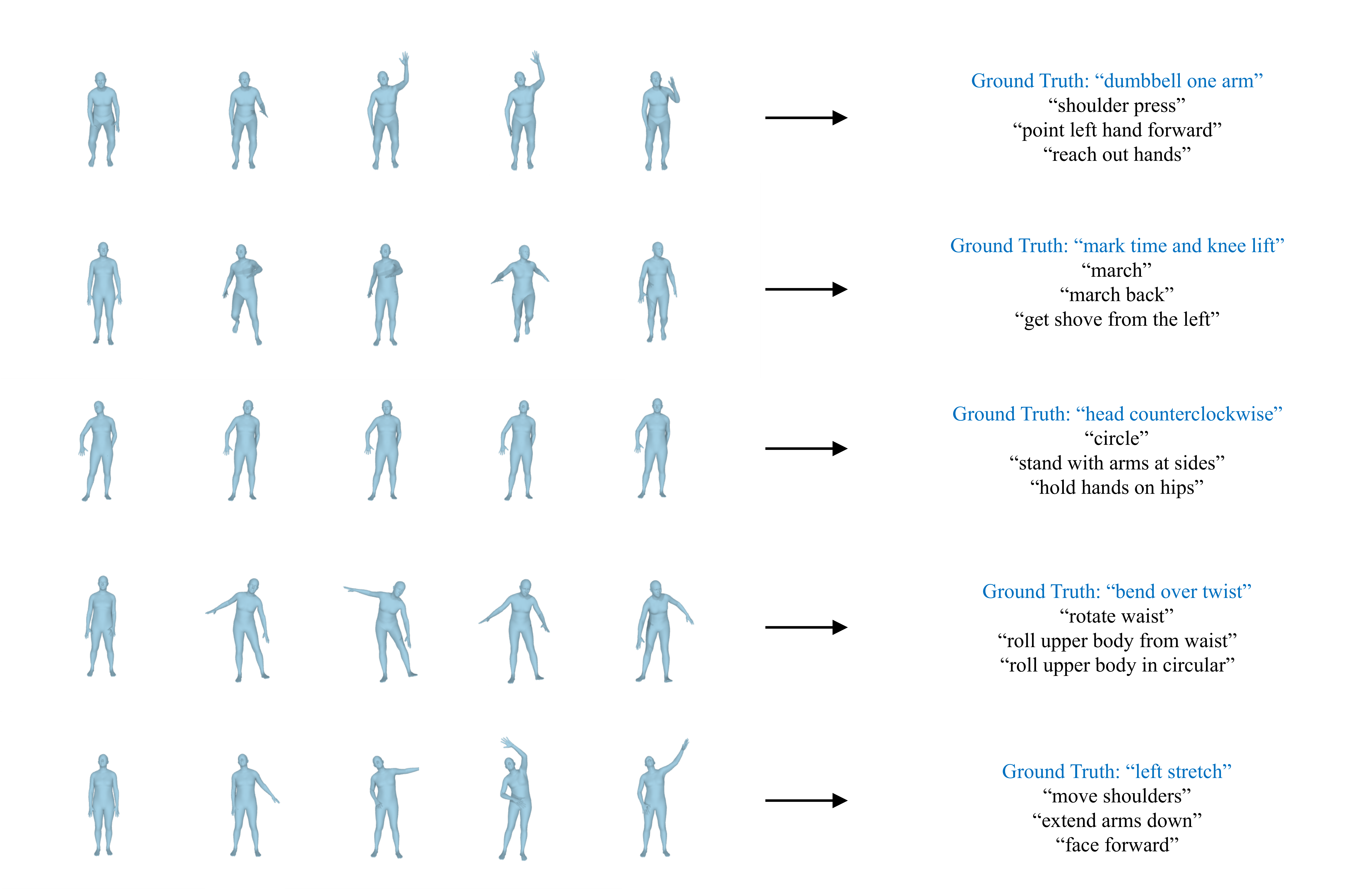}
\caption{Action recognition for UESTC dataset. Top-3 predicted labels from BABEL dataset.}
\label{fig:app_action_recognition_uestc}
\end{figure*}

More action recognition examples in four datasets are provided in Figure~\ref{fig:app_action_recognition_babel}, Figure~\ref{fig:app_action_recognition_humanact12}, Figure~\ref{fig:app_action_recognition_nturgbd13}, and Figure~\ref{fig:app_action_recognition_uestc}.

\subsection{More Motion Retrieval Results}
\label{supp:moreretrieval}

\begin{figure*}[!ht]
\centering
\includegraphics[width=\textwidth]{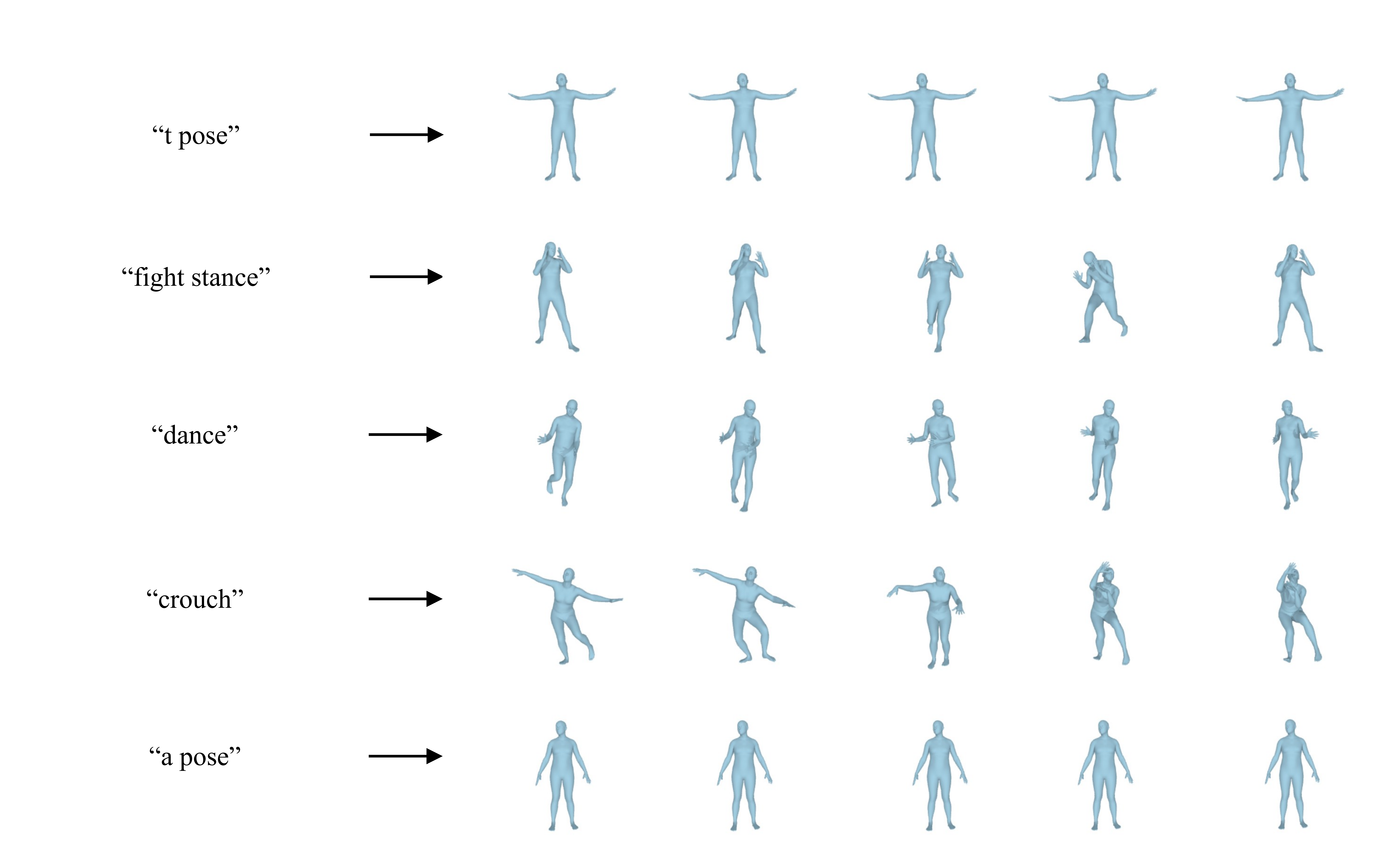}
\caption{Motion retrieval from BABEL dataset.}
\label{fig:app_mr_babel}

\end{figure*}

An additional motion retrieval example is provided in Figure~\ref{fig:app_mr_babel}.

\section{Appendix 2: t-SNE Visualization of Language Representations}
\label{supp:sec_with_lang_repr}

\begin{figure*}[!ht]
\centering
\includegraphics[width=\textwidth]{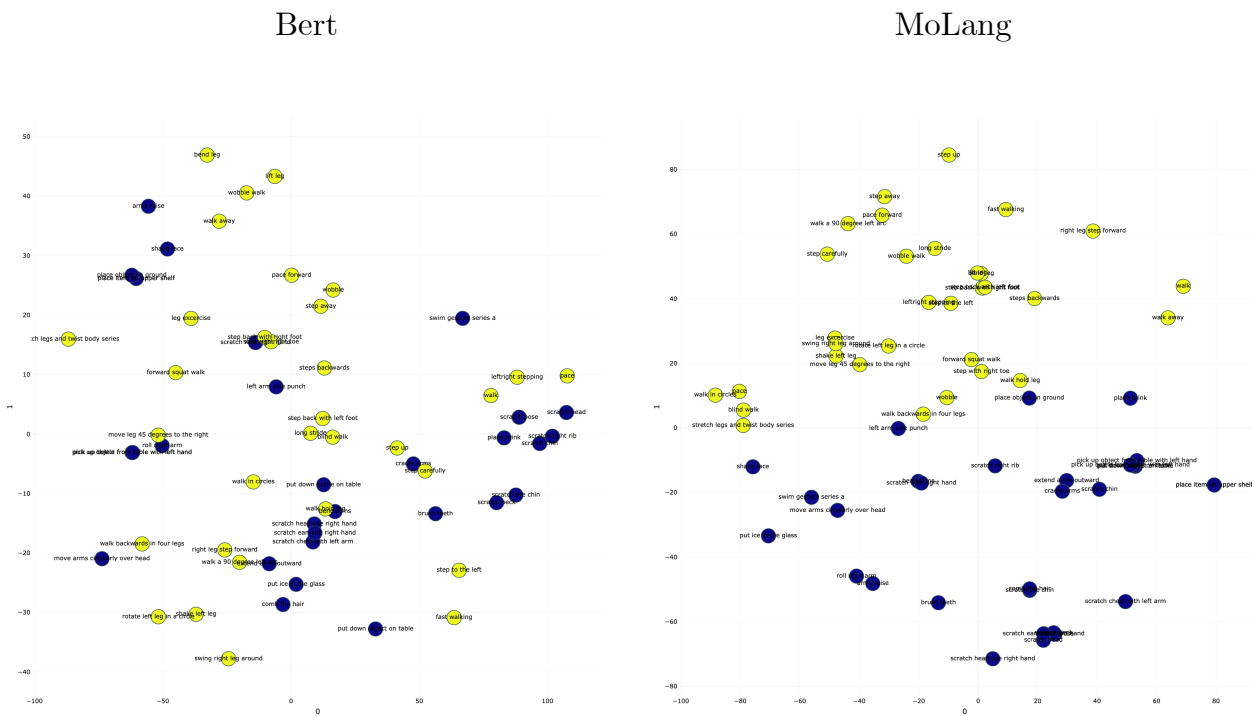}
\caption{t-SNE results from text encoder. \textbf{(yellow)}: motion using legs. \textbf{(blue)}: motion using arms.}
\label{fig:tsne}
\end{figure*}

\begin{figure*}[!ht]
\centering
\includegraphics[width=\textwidth]{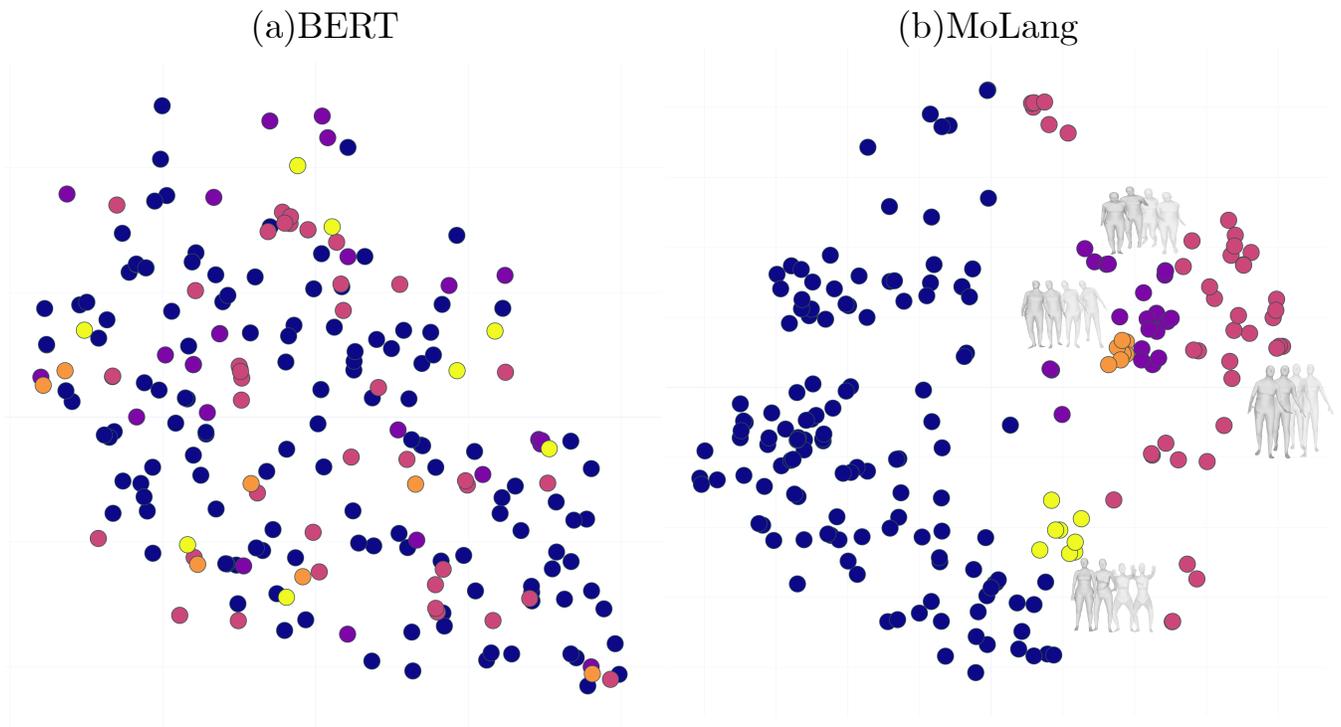}
\caption{t-SNE results from text encoder. \textbf{(yellow)}: motion using knees. \textbf{(pink)}: walking motion. \textbf{(orange)}: body rotating motion. \textbf{(purple)}: jumping motion.}
\label{fig:tsne-leg}
\end{figure*}

\begin{figure*}[ht]
\centering
\includegraphics[width=\textwidth]{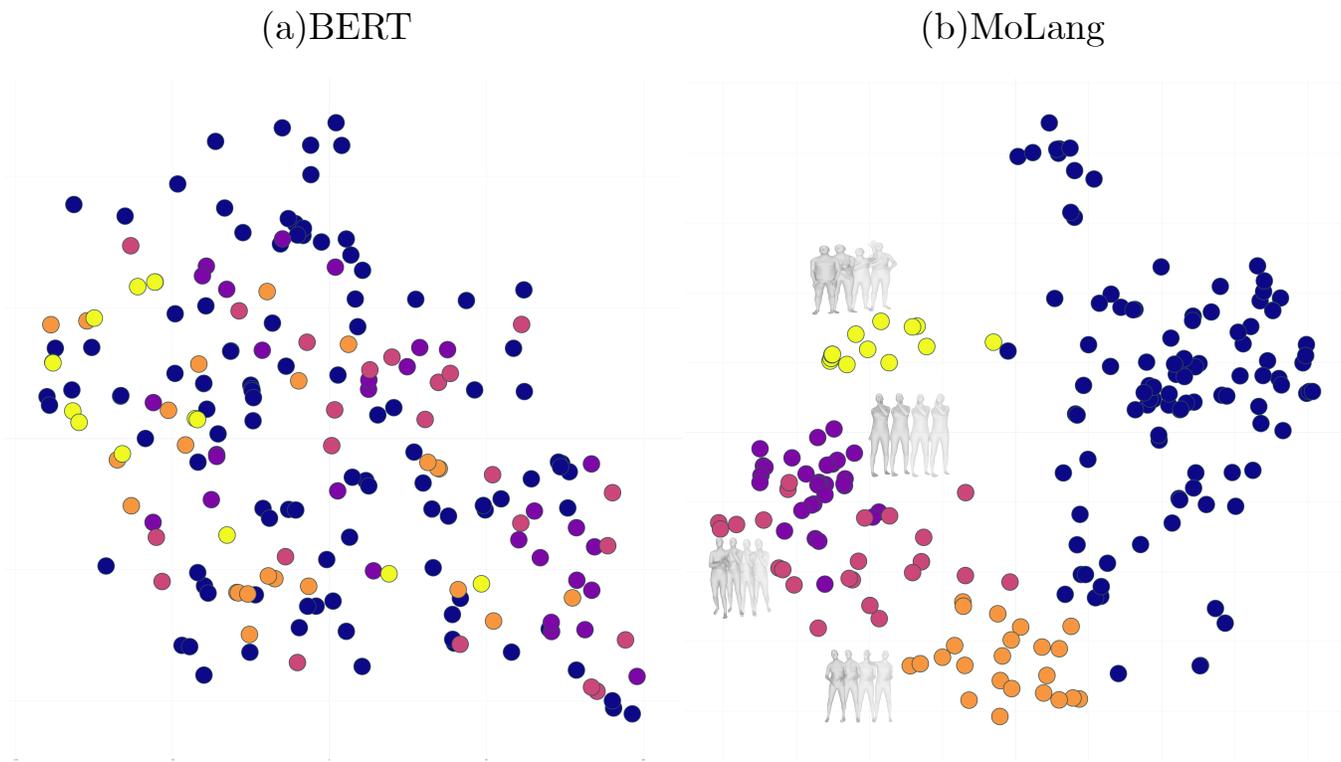}
\caption{t-SNE results from text encoder. \textbf{(yellow)}: throwing motion. \textbf{(pink)}: motion that move arms near chest. \textbf{(orange)}: object interaction using hands. \textbf{(purple)}: motion of moving arms near head.}
\label{fig:tsne-arm}
\end{figure*}

\begin{figure*}[!ht]
\vspace{-10mm}
\centering
\includegraphics[width=\textwidth]{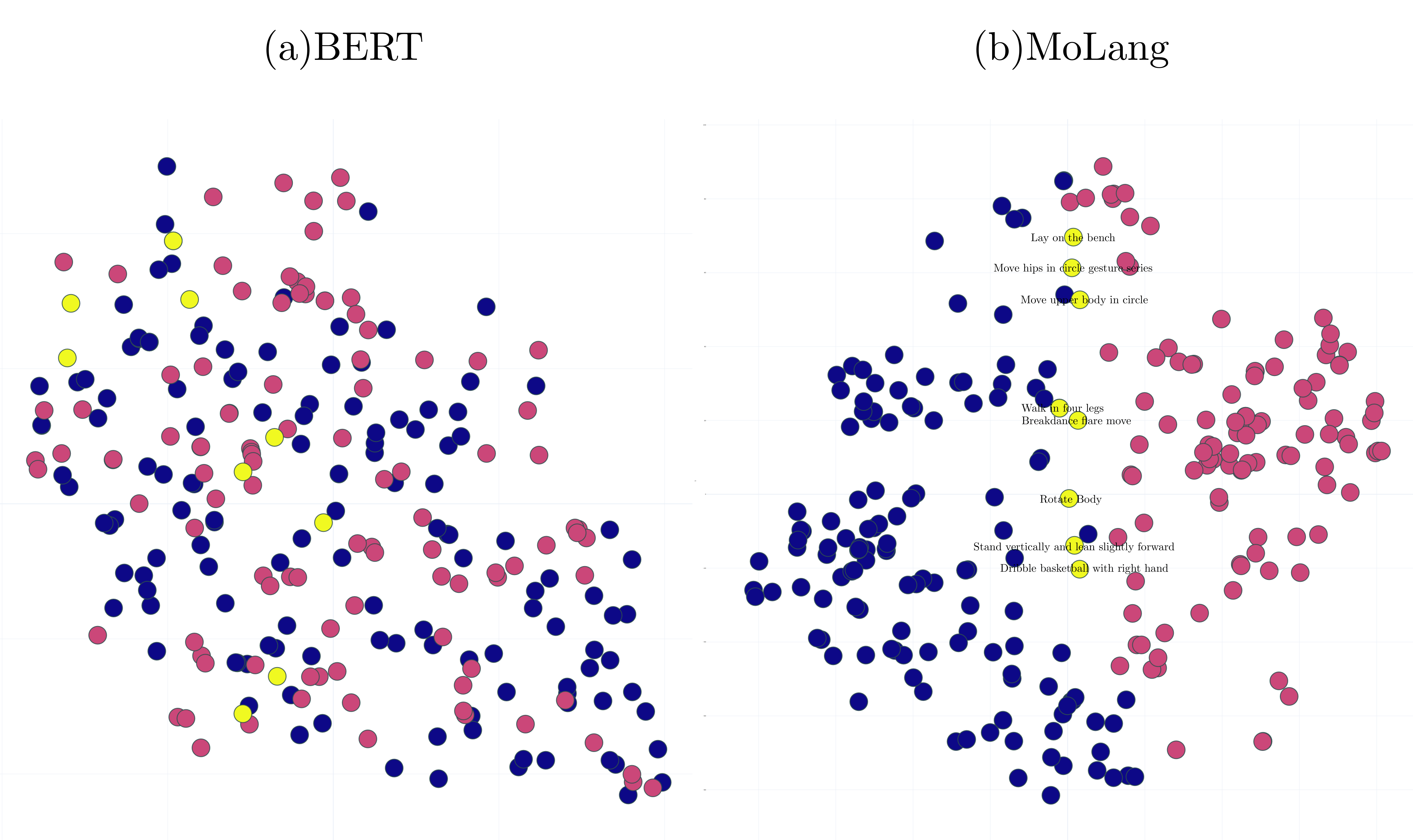}
\caption{t-SNE results from text encoder. \textbf{(yellow)}: motion using legs. \textbf{(blue)}: motion using arms. \textbf{(pink)}: motion using legs.}
\label{fig:tsne-between}
\end{figure*}

Visualizations of latent space of text encoder in t-SNE. Figure~\ref{fig:tsne}, Figure~\ref{fig:tsne-leg}, Figure~\ref{fig:tsne-arm}, and Figure~\ref{fig:tsne-between} show better understanding of the relation between motion and text rather than the original text encoder.

\section{Appendix 3: Hyperparameters Experimental Setup }
\label{supp:hyperparameters}
Hyperparameters of motion encoder and MoLang are listed in Table~\ref{tab:hyperp_mmp} and Table~\ref{tab:hyperp_molang}.

\begin{table}[h]
    \centering
    \begin{tabular}{c|c} \hline
        Hyperparameter & Motion Encoder  \\ \hline
        Optimizer & Adam \\
        Learning Rate & 0.0001 \\
        $\beta_1$ & 0.9 \\
        $\beta_2$ & 0.999 \\
        \hline
        Scheduler & \textrm{Cosine Annealing} \\
        $T_0$ & 20 \\
        $T_{\textrm{mult}}$  & 1 \\
        $\eta_{\textrm{min}}$ & 0.00007 \\
        \hline
        Batch Size & 512 \\
        Num Layers & 10 \\ 
        Heads & 12 \\
        Maximum Length & 150 \\
        Feedforward Dimension & 1024 \\
        Dropout Rate & 0.1 \\
        Embedding Dimension & 768 \\
        GCB Layer & \textrm{between 4th and 5th Layer}\\ \hline
    \end{tabular}
    \caption{Hyperparameters for training motion encoder}
    \label{tab:hyperp_mmp}
\end{table}

\begin{table}[h]
    \centering
    \begin{tabular}{c|c} \hline
        Hyperparameter & MoLang  \\ \hline
        Optimizer & Adam \\
        Learning Rate & 0.0001 \\
        $\beta_1$ & 0.9 \\
        $\beta_2$ & 0.999 \\
        \hline
        Scheduler & \textrm{Cosine Annealing} \\
        $T_0$ & 20 \\
        $T_{\textrm{mult}}$  & 1 \\
        $\eta_{\textrm{min}}$ & 0.00007 \\
        \hline
        Text Encoder & Pretrained BERT (`bert-base-uncased`) \\ \hline
        Temperature & 0.07 \\ 
        Weight for $\mathcal{L}_{\textrm{recon}}$ & 10 \\
        \hline
    \end{tabular}
    \caption{Hyperparameters for training MoLang}
    \label{tab:hyperp_molang}
\end{table}

\end{document}